\documentclass[a4paper,conference]{IEEEtran}
\usepackage{graphicx}
\usepackage{amsmath}
\usepackage{multirow}
\usepackage{cite}

\ifCLASSINFOpdf
\else
\fi
\begin{document}
%
\title{Boosting Video Captioning with Dynamic Loss Network}

\author{\IEEEauthorblockN{Nasib Ullah}
\IEEEauthorblockA{Electronics and Communication Sciences Unit\\
Indian Statistical Institute\\
Kolkata, India\\
Email: nasibullah104@gmail.com}
\and
\IEEEauthorblockN{Partha Pratim Mohanta}
\IEEEauthorblockA{Electronics and Communication Sciences Unit\\
Indian Statistical Institute\\
Kolkata, India \\
Email: ppmohanta@isical.ac.in}}


%


\maketitle

\begin{abstract}
Video captioning is one of the challenging problems at the intersection of vision and language, having many real-life applications in video retrieval, video surveillance, assisting visually challenged people, Human-machine interface, and many more. Recent deep learning based methods \cite{s2vt_ref,marn_ref,orgtrl_ref}  have shown promising results but are still on the lower side than other vision tasks (such as image classification, object detection). A significant drawback with existing video captioning methods is that they are optimized over cross-entropy loss function, which is uncorrelated to the de facto evaluation metrics (BLEU, METEOR, CIDER, ROUGE). In other words, cross-entropy is not a proper surrogate of the true loss function for video captioning. To mitigate this, methods like REINFORCE, Actor-Critic, and Minimum Risk Training (MRT) have been applied but have limitations and are not very effective. This paper proposes an alternate solution by introducing a dynamic loss network (DLN), providing an additional feedback signal that reflects the evaluation metrics directly. Our solution proves to be more efficient than other solutions and can be easily adapted to similar tasks. Our results on Microsoft Research Video Description Corpus (MSVD) and MSR-Video to Text (MSRVTT) datasets outperform previous methods.
\end{abstract}


%
\IEEEpeerreviewmaketitle

\section{Introduction}
Video captioning is the task of describing the content in a video in natural language. With the explosion of sensors and the internet as a data carrier, automatic video understanding and captioning have become essential. It can be applied in many applications such as video surveillance, assisting visually challenged people, video retrieval, and many more. Despite having many applications, jointly modeling the spatial appearance and temporal dynamics makes it a difficult task. \\

 Motivated by machine translation \cite{mt2_ref} and image captioning \cite{imgcap1_ref,imgcap2_ref}, the encoder-decoder architecture has been adapted for the video captioning task \cite{meanpool_ref,salstm_ref,marn_ref,recnet_ref,s2vt_ref}. On the encoder side, different visual features are extracted using 2D and 3D convnets. The encoder's combined visual features are sent to the decoder to generate the caption, one word at a time. So basically, the decoder is a conditional language model, and a variant of recurrent neural networks (LSTM, GRU) is the most popular and successful. Recent improvements on the encoder-decoder baseline have happened in mainly three areas: (i) incorporation of better visual feature extraction modules at the encoder side, (ii) addition of external language models to guide the decoder, (iii) better frame selection strategy. 
\begin{figure}[t]
\centering
\includegraphics[width=\columnwidth]{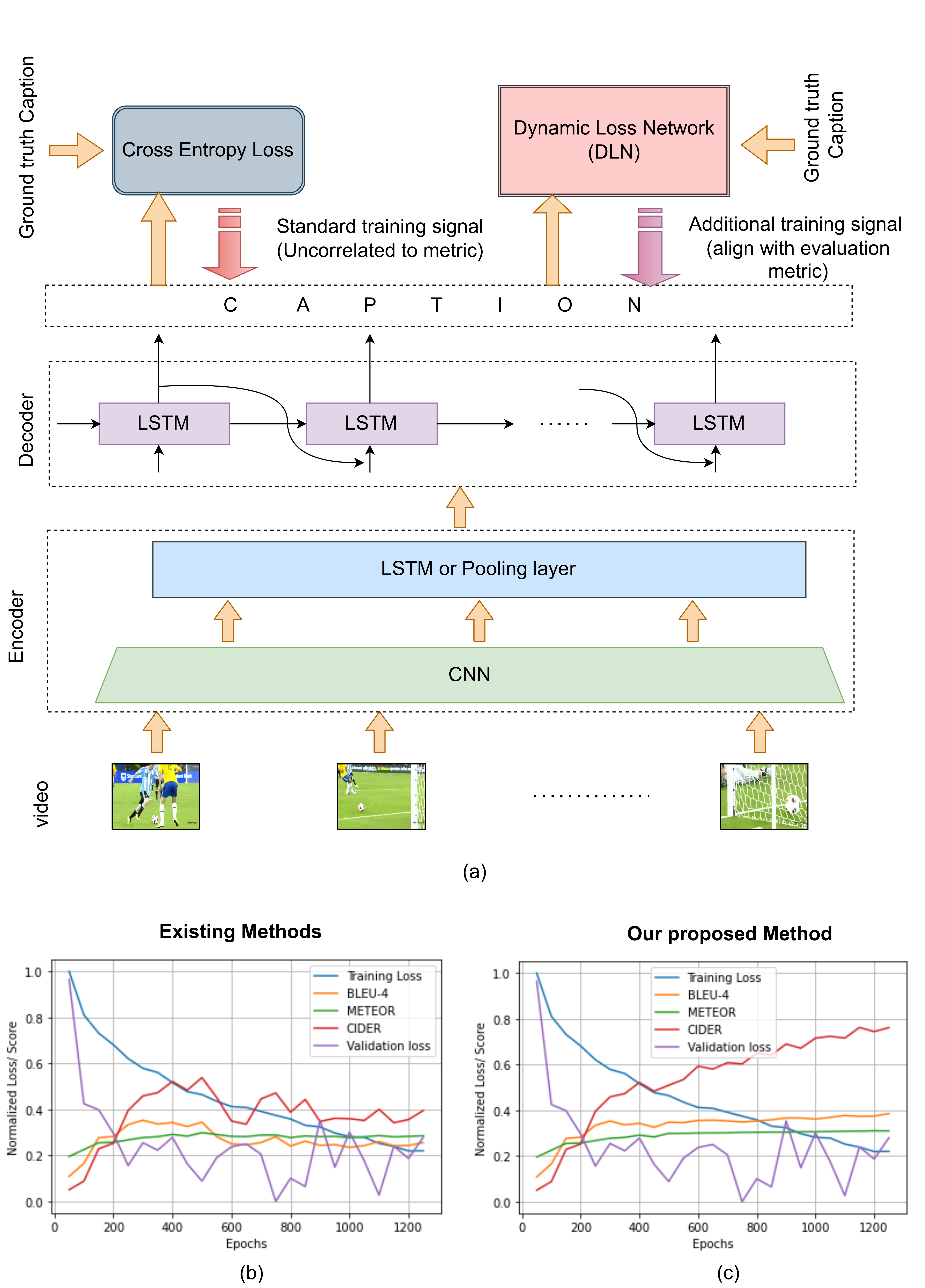}
\caption{(a) The proposed Dynamic loss network (DLN) with an encoder-decoder architecture. The encoder-decoder relies on the standard cross-entropy training signal whereas the DLN introduces additional training signal aligned with the evaluation metrics. (b) Training signal and evaluation metric curve for a standard encoder-decoder architecture. (c) Training signal and evaluation metric curve for our proposed architecture.}
\label{fig1}
\end{figure}
Despite the improvements, a potential drawback with these methods is that the training signal does not align with the standard evaluation metrics such as BLEU \cite{bleu_ref}, METEOR \cite{meteor_ref}, ROUGE-L \cite{rougel_ref}, CIDER \cite{cider_ref}. As a result, even low training and validation loss can lead to poor metric scores and vice versa, as shown in Fig.\ref{fig1}(b).  Furthermore, direct optimization over metric function is not possible due to the non-differentiable nature of the network. Alternate solutions from Reinforcement learning (REINFORCE, Actor-Critic) and Minimum Risk Training (MRT) have been applied to machine translation and image captioning. However, they have not proved to be very successful in the case of video captioning. To this end, we propose a dynamic loss network (DLN), a transformer-based model that approximates metric function and is pre-trained on external data using a self-supervised setup. Although the proposed DLN can be utilized to approximate any metric function, in our case, we approximate the BLEU, METEOR, and CIDER scores. Once trained, the DLN can be used with the video captioning model in an end-to-end manner, as shown in Fig.\ref{fig1}(a). 

Finally, we demonstrate that the feedback signals from our proposed model align with the evaluation metric, as shown in Fig.\ref{fig1}(c).

\section{Related Work}
\subsection{Video Captioning.}The main breakthrough in video captioning happened with the inception of encoder-decoder based sequence to sequence models. The encoder-decoder framework for video captioning was first introduced by MP-LSTM \cite{meanpool_ref}, which uses mean pooling over-frame features and then decodes caption by LSTM. Although MP-LSTM \cite{meanpool_ref} outperformed its predecessors, the temporal nature of the video was first modeled by S2VT \cite{s2vt_ref} and SA-LSTM \cite{salstm_ref}. The former shares a single LSTM for both the encoder and the decoder, while the latter uses attention over-frame features along with 3D HOG features. The recent methods are improved on the SA-LSTM \cite{salstm_ref} baseline. RecNet \cite{recnet_ref} uses backward flow and reconstruction loss to capture better semantics, whereas MARN \cite{marn_ref} uses memory to capture correspondence between a word and its various similar visual context. M3 \cite{m3_ref} also uses memory to capture long-term visual-text dependency, but unlike MARN \cite{marn_ref}, it uses heterogeneous memory. Both MARN \cite{marn_ref} and M3 \cite{m3_ref} use motion features along with appearance features. More recently, STG-KD \cite{stgkd_ref} and OA-BTG \cite{oabtg_ref} use object features along with the appearance and motion features. STG-KD \cite{stgkd_ref} uses a Spatio-temporal graph network to extract object interaction features, whereas OA-BTG \cite{oabtg_ref} uses trajectory features on salient objects. ORG-TRL \cite{orgtrl_ref} uses Graph convolutional network (GCN) to model object-relational features and an external language model to guide the decoder. Another group of methods focuses on devising a better sampling strategy to pick informative video frames. PickNet \cite{picknet_ref} uses reward-based objectives to sample informative frames, whereas SGN \cite{sgn_ref} uses partially decoded caption information to sample frames. Despite the improvements, all these methods suffer from improper training signals, and some effort has already been made to mitigate this issue.
\subsection{Training on evaluation metric function.} There are mainly three approaches to optimize the sequence to sequence model on the non-differentiable objective function: (i) Ranzato et al. \cite{reinforce_ref} use the REINFORCE algorithm \cite{reinforce_original_ref} to train an image captioning model directly on BLEU score and Rennie et al. \cite{selfcritical_ref} use the Actor-critic method \cite{actorcritic_ref}. Both methods use the reward signal, but these methods are not applicable for video captioning due to the sparse nature of the reward. (ii) Optimization on differentiable lower bound where Zhukov et al. \cite{lowerboundbleu_ref} propose a differentiable lower bound of expected BLEU score and Casas et al. \cite{fonollosa_ref} reported poor training signal corresponding to their formulation of differentiable BLEU score \cite{bleu_ref}. (iii) Shiqi Shen et al. \cite{mrt_ref} use Minimum risk training (MRT) instead of Maximum likelihood estimation  for neural machine translation, and Wang et al. \cite{mrt2_ref} shows Minimum Risk Training (MRT) helps in reducing exposure bias. Unlike previous works, we leverage successful Transformer based pre-trained models to approximate the evaluation metrics.

\section{Method}
\begin{figure*}[t]
\includegraphics[width=\textwidth,height=360px]{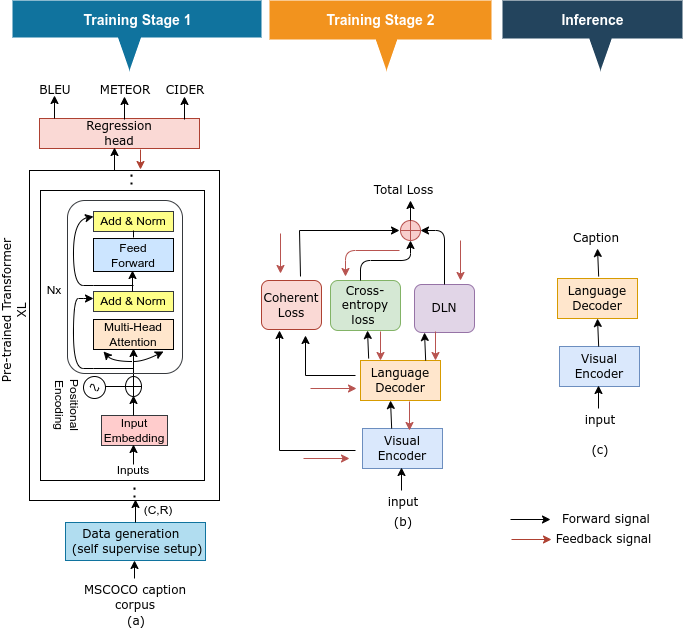}
\caption{(a) Training of Dynamic Loss Network in self-supervised setup. (b) End-to-end training of video captioning model along with DLN. (c) Video captioning model at test time.}\label{fig2}
\end{figure*}

Our proposed method follows a two-stage training process. At the first stage, the DLN is trained in a self-supervised setup, whereas at the second stage, the trained DLN is used along with the existing video captioning model. The entire process flow is in the Fig.\ref{fig2}. During the second stage, the loss from the DLN back propagates through the encoder-decoder model and forces it to capture better representation. Moreover, the proposed loss network can be combined with different encoder-decoder architectures for video captioning. Below we describe each component of our model.

\subsection{Visual Encoder}
We uniformly sample N frames $\{f_{i}\}_{i=1}^{N}$ and clips $\{c_{i}\}_{i=1}^{N}$ from a given input video, where each $c_{i}$ is a series of clips surrounding frame $f_{i}$. We extract appearance features $\{a_{i}\}_{i=1}^{N}$ and motion features $\{m_{i}\}_{i=1}^{N}$ using pre-trained 2D convnets \cite{vit_ref} $\Phi^{a}$  and 3D convnets \cite{c3d_ref} $\Phi^{m}$ , with $a_{i}= \Phi^{a}(f_{i})$ and $m_{i}= \Phi^{m}(c_{i})$, respectively. Apart from appearance $(\{a_{i}\}_{i=1}^{N})$ and motion $(\{m_{i}\}_{i=1}^{N})$, we extract object characteristics $(\{o_{i}\}_{i=1}^{N})$ through a pre-trained object detection module $\Phi^{o}$, where $o_{i}= \Phi^{o}(f_{i})$. We select prominent items from each frame based on the objectiveness threshold v and average their features. The appearance and motion characteristics aid in comprehending the video's global context and motion information. By contrast, object characteristics are more localized, which aids in the comprehension of fine-grained information.

\subsection{Dynamic Loss Network (DLN)} 
As shown in Fig.\ref{fig1}(a), the proposed DLN is built on top of the encoder-decoder and provides an additional training signal aligned with the evaluation metric. The proposed DLN approximates the evaluation metric BLEU \cite{bleu_ref}, METEOR \cite{meteor_ref}, and CIDER \cite{cider_ref}, which involves mapping from a pair of sentences to numerical values. Motivated by the tremendous success in vision and natural language processing (NLP), a pre-trained transformer network \cite{transformer_ref,bert_ref,gpt_ref,transformerxl_ref} is used as the backbone for the proposed DLN.

The training of the DLN is achieved in a self-supervised manner. The training data and its ground truth to train the DLN (Fig. \ref{fig2}(a)) are generated following two strategies: (i) we take MSCOCO \cite{mscoco_ref} caption corpus and perturb each sentence randomly with a p\% probability to generate (candidate $C_{i}$, reference $R_{i}$) pair. For the perturbation, deletion and swapping are done over the word(s). (ii) we train a standard encoder-decoder based video captioning model and gather the predicted and ground truth caption as (candidate, reference) pair at different epochs on MSVD \cite{msvd_ref} data. In both cases, ground truth (BLEU, METEOR, and CIDER) is generated using the NLTK \cite{nltk_ref} library and the COCO evaluation \cite{cocoeval_ref} server. 

The self-attention layer in the transformer network (to be more specific, transformer network with the word as input) calculates the attention score between words. This characteristic makes the transformer network \cite{transformer_ref} a natural choice to model the metric score function (since BLEU, METEOR, and CIDER are precision and recall based formulas on the n-gram overlap). Although BERT \cite{bert_ref} and GPT \cite{gpt_ref} are state-of-the-art pre-trained transformer architecture, they are not suitable to model metric scores due to subword input tokenization. Instead, we use TransformerXL \cite{transformerxl_ref} architecture, which works with standard word input (similar to the LSTM decoder). A regression head has been added on top of the standard TransformerXL \cite{transformerxl_ref} network and trained by minimizing the mean square loss between the true and predicted BLEU, METEOR, and CIDER values. The output of DLN is,
\begin{equation}
    t_{i} = W \Upsilon(C_{i},R_{i}) + b
\end{equation}
where, $t_{i}$ = $(t_{i}^{BLEU},t_{i}^{METEOR},t_{i}^{CIDER})$, $\Upsilon$ is transformerXL model, W  and b are the learnable parameters corresponding to regression head. R, C are reference and candidate sentences, respectively.

Once trained, the DLN is combined with the standard encoder-decoder network at the second stage of training. The proposed DLN is applied only at the training stage, so there is no run-time overhead during inference. As shown in Fig.\ref{fig2}(b), the DLN takes inputs from the output of the decoder and ground truth caption. During the backward pass, the output value of DLN is added to cross-entropy loss, and the model is trained on the combined loss function. 

\subsection{Language Decoder} The decoder generates the caption word by word based on the features obtained from the visual encoder. A recurrent neural network is utilized as the backbone of the decoder because of its superior temporal modeling capability. In the proposed system, the decoder is designed using LSTM \cite{meanpool_ref}, whose hidden memory at time step $t$ can be expressed as
\begin{equation}
    h_{t} = LSTM(C_{t},h_{t-1})
\end{equation}
Where $C_{t}$ is the concatenation of appearance, motion, and object features from the visual encoder and $h_{t-1}$ is the hidden memory of time step $t-1$. To predict the word probability, a linear layer followed by a Softmax layer is added on top of the hidden layers of the LSTM.

\begin{equation}
    P(s_{t}|V,s_{1},s_{2},..,s_{t-1}) = Softmax(V_{h}h_{t}+b_{h})
\end{equation}
 where $s_{t}$ is the $t^{th}$ word in the caption and $V_{h}$ and $b_{h}$ are the learnable parameters and biases, respectively.

\subsection{Parameter Learning}Along with the typical cross-entropy loss, we train our model with two extra losses: Loss from DLN and Coherent loss.
\subsubsection{Language Decoder} The cross-entropy or negative log-likelihood function is the typical loss function for an encoder-decoder based video captioning model. For a mini-batch, the loss can be expressed as
\begin{equation}
    \textit{L}_{LD} = - \sum_{i=1}^B \sum_{t=1}^T \log p(s_{t}|V,s_{1},s_{2},..,s_{t-1};\theta)
\end{equation}
Where $\theta$ is learnable parameters, $V$ is the video feature, $s_{t}$ is the $t^{th}$ word in the sentence of length T, and B is the mini-batch size. 
\subsubsection{DLN Loss} The proposed DLN works in two stages. We train the DLN to predict BLEU, METEOR, and CIDER scores first. We use the Mean square error loss function as the objective for this task, and for a mini-batch, it can be expressed as,
\begin{equation}
\begin{aligned}
    \textit{L}_{DLN}^{1} = & \sum_{i=1}^B [\lambda_{1}^{1} (y_{i}^{BLEU} - t_{i}^{BLEU}) \\
                        & + \lambda_{2}^{1} (y_{i}^{METEOR} - t_{i}^{METEOR}) \\
                        & + \lambda_{3}^{1} (y_{i}^{CIDER} - t_{i}^{CIDER}) ]
\end{aligned}
\end{equation}
where, $y_{i}^{}$ is the ground truth and $t_{i}^{}$ is the model prediction. $\lambda_{1}^{1}$, $\lambda_{2}^{1}$, and $\lambda_{3}^{1}$ are hyperparameters to control the relative imporance of three different losses.

The DLN predicts BLEU, METEOR, and CIDER score at the second stage and uses it to optimize the encoder-decoder model. For a mini-batch, the loss is
\begin{equation}
\begin{aligned}
    \textit{L}_{DLN} = & -\sum_{i=1}^B[\lambda_{BLEU} t_{i}^{BLEU} + \lambda_{METEOR} t_{i}^{METEOR} \\
                    & + \lambda_{CIDER} t_{i}^{CIDER}]
\end{aligned}
\end{equation}
where, $t_{i}^{BLEU}$, $t_{i}^{METEOR}$, $t_{i}^{CIDER}$ are the predicted BLEU, METEOR and CIDER scores from the DLN respectively and $\lambda_{BLEU}$, $\lambda_{METEOR}$ and $\lambda_{CIDER}$ are the hyperparameters.
\subsubsection{Coherent Loss} A video's successive frames are exceedingly repetitious. As a result, the encoding of subsequent frames should be comparable. We use the coherence loss to constrain subsequent frames' embeddings to be comparable. Coherent loss has been used before to normalise attention weights \cite{marn_ref}; however, unlike Pei at al. \cite{marn_ref}, we use the coherent loss to appearance, motion, and object aspects. For a mini-batch, the total coherence loss is,
\begin{equation}
    \textit{L}_{C} = \lambda_{fc} \textit{L}_{C}^a + \lambda_{mc} \textit{L}_{C}^m + \lambda_{oc} \textit{L}_{C}^o + \lambda_{ac} \textit{L}_{C}^{\alpha}  
\end{equation}
where $\lambda_{fc}$, $\lambda_{mc}$, $\lambda_{oc}$ and $\lambda_{ac}$ are hyperparameters corresponding to appearance coherent loss $\textit{L}_{C}^a$, motion coherent loss $\textit{L}_{C}^m$, object coherent loss $\textit{L}_{C}^o$ and attention coherent loss $\textit{L}_{C}^\alpha$ respectively.

\noindent The individual coherent losses are calculated as, $\textit{L}_{C}^a=\Psi(a_{i}^r)$,     $\textit{L}_{C}^m =\Psi(m_{i}^r)$, $\textit{L}_{C}^o=\Psi(o_{i}^r)$ and           $\textit{L}_{C}^{\alpha}=\Psi(\alpha_{i})$ where,
\begin{equation}
    \Psi(f) = \sum_{i=1}^B \sum_{t=1}^T \sum_{n=2}^N | f_{n,t}^{(i)} - f_{n-1,t}^{(i)} |
\end{equation}

\begin{table*}[t]
    \centering
    \resizebox{0.95\textwidth}{!}{
    \begin{tabular}{c|c c c c|c c c c}
      \hline \hline 
     Models &   & MSVD &  &  &  & MSR-VTT &  &  \\
     & B@4  & M & R & C & B@4 & M & R & C \\ 
     \hline 
     SA-LSTM \cite{salstm_ref} & 45.3 & 31.9 & 64.2 &76.2 & 36.3 & 25.5 & 58.3 & 39.9 \\
     h-RNN \cite{hrnn_ref} & 44.3 & 31.1 & - & 62.1 & - & - & - & - \\
     hLSTMat \cite{hlstmatt_ref} &53.0 & 33.6 & - & 73.8 & 38.3 & 26.3 & - & - \\ 
     RecNet \cite{recnet_ref} & 52.3 & 34.1 & 69.8 & 80.3 & 39.1 &26.6 & 59.3 &42.7 \\
     M3 \cite{m3_ref} & 52.8 & 33.3 & - & - & 38.1 & 26.6 & - & - \\
     PickNet \cite{picknet_ref} & 52.3 & 33.3 & 69.6 & 76.5 & 41.3 & 27.7 & 59.8 & 44.1 \\
     MARN \cite{marn_ref} & 48.6 & 35.1 & 71.9 & 92.2 & 40.4 & 28.1 & 60.7 & 47.1 \\
     GRU-EVE \cite{grueve_ref} & 47.9 & 35.0 & 71.5 & 78.1 & 38.3 & 28.4 & 60.7 & 48.1 \\
     POS+CG \cite{poscg_ref} & 52.5 & 34.1 & 71.3 & 88.7 & 42.0 & 28.2 & 61.6 & 48.7 \\
     OA-BTG \cite{oabtg_ref} & \textbf{56.9} & 36.2 & - & 90.6 & 41.4 & 28.2 & - & 46.9 \\
     STG-KD \cite{stgkd_ref} & 52.2 & \textbf{36.9} & 73.9 & 93.0 & 40.5 & 28.3 & 60.9 & 47.1 \\ 
     SAAT \cite{saat_ref} & 46.5 & 33.5 & 69.4 & 81.0 & 40.5 & 28.2 & 60.9 & 49.1 \\
     ORG-TRL \cite{orgtrl_ref} & 54.3 & 36.4 & 73.9 & 95.2 & \textbf{43.6} & 28.8 & \textbf{62.1} & 50.9 \\
     SGN \cite{sgn_ref} & 52.8 & 35.5 & 72.9 & 94.3 & 40.8 & 28.3 & 60.8 & 49.5 \\
     \hline  
     Ours & 53.1 & 36.3 & \textbf{74.1} & \textbf{97.4} & 41.3 & \textbf{29.1} & 61.8 & \textbf{51.5} \\
     \hline \hline \\
    \end{tabular}}
    \caption{Performance comparison on MSVD and MSR-VTT benchmarks. B4, M, R, and C denote BLEU{-}4, METEOR, ROUGE\_L, and CIDER, respectively.}
    \label{tab:main_table}
\end{table*}

At the early training phase, cross entropy acts as a better training signal, so we rely more on cross entropy loss. On the other hand, we rely more on loss from the proposed loss network at the later phase of training. The total loss for a mini-batch is
\begin{equation}
    \textit{L} = \textit{L}_{LD} + \textit{L}_{DLN} + \textit{L}_{C}
\end{equation}

\section{Experiments and Results}
We have conducted experiments to evaluate the proposed DLN-based video captioning performance on two benchmark datasets: Microsoft Research-Video to Text (MSRVTT) \cite{msrvtt_ref} and Microsoft Research Video Description Corpus (MSVD) \cite{msvd_ref}. In addition, We have compared the performance of our method with the state-of-the-art video captioning methods. Adding DLN provided significant gain to the captioning performance in all metrics.

\subsection{Datasets}
\subsubsection{MSVD} MSVD contains open domain 1970 Youtube videos with approximately 40 sentences per clip. Each clip contains a single activity in 10 seconds to 25 seconds. We have followed the standard split \cite{meanpool_ref,salstm_ref,marn_ref} of 1200 videos for training, 100 for validation,  and 670 for testing. 
\subsubsection{MSRVTT} MSRVTT is the largest open domain video captioning dataset with 10k videos and 20 categories. Each video clip is annotated with 20 sentences, resulting in 200k video-sentence pairs. We have followed the public benchmark splits, i.e., 6513 for training, 497 for validation, and 2990 for testing.
\subsection{Implementation Details}
We have uniformly sampled 28 frames per video and extracted 1024D appearance features from Vision Transformer \cite{vit_ref}, pre-trained on ImageNet \cite{imagenet_ref}. The motion features are 2048D and extracted using C3D \cite{c3d_ref} with ResNeXt-101 \cite{resnext_ref} backbone and pre-trained on Kinetics-400 dataset. We use Faster-RCNN \cite{fasterrcnn_ref} pre-trained on MSCOCO \cite{mscoco_ref} for object feature extraction. Appearance,motion, and object features are projected to 512D before sending to the decoder. At the decoder end, the hidden layer and the size of the word embedding are both set as 512D. The dimension of the attention module is set to 128D. All the sentences longer than 30 words are truncated, and the vocabulary is built by words with at least 5 occurrences. For the DLN, we use 16 multi-head and 18 layers TransformerXL \cite{transformerxl_ref} pre-trained on WikiText-103. A regression head composed of three fully connected (FC) layers is added on the top of the TransformerXL \cite{transformerxl_ref}. During both stages of training, the learning rate for DLN and the end-to-end video captioning model is set to 1e-4. Adam \cite{adam_ref} is employed for optimization. The model selection is made using the validation set performance. The greedy search is used for the caption generation at the test time. The coherent loss weights $\lambda_{ac}$, $\lambda_{fc}$, $\lambda_{mc}$, and $\lambda_{oc}$ are set as 0.01, 0.1, 0.01, and 0.1, respectively. All the experiments are done in a single Titan X GPU.

\subsection{Quantitative Results} We have compared our proposed model with the existing video captioning models on MSVD and MSRVTT datasets, as shown in Table \ref{tab:main_table}. All four popular evaluation metrics, including BLEU, METEOR, ROUGE, and CIDER, are reported. From Table \ref{tab:main_table}, we can see that our proposed method significantly outperforms other methods, especially in the CIDER score. It is to be noted that CIDER is specially designed to evaluate captioning tasks. Compared to current methods (ORG-TRL \cite{orgtrl_ref}, STG-KD \cite{stgkd_ref}, SAAT \cite{saat_ref}), which uses more complex object-relational features, our method only takes mean object localization features for simplicity and to prove the effectiveness of the DLN.

\begin{table}[h]
    \centering
    \resizebox{\columnwidth}{!}{
    \begin{tabular}{c|c c|c c}
      \hline \hline 
     Models & Without DLN &  & With DLN &    \\
     &  M & C  & M & C \\ 
     \hline 
     SA-LSTM \cite{salstm_ref} & 31.9 & 76.2 & 33.1 & 77.2 \\
     RecNet \cite{recnet_ref} & 34.1 & 80.3 & 34.4 & 81.3  \\
     M3 \cite{m3_ref} & 33.3 & - & 34.9 & -  \\
     PickNet \cite{picknet_ref} & 33.3 & 76.5 & 34.5 & 78.7  \\
     MARN \cite{marn_ref} & 35.1 & 92.2 & 35.7 & 93.4  \\
     \hline \hline
    \end{tabular}}
    \caption{Ablation studies on  MSVD benchmark. M and C denote METEOR and CIDER respectively.}
    \label{tab:abl_1}
\end{table}

\subsection{Ablation Studies } In order to validate the effectiveness of the proposed DLN and prove that improvement is not because of the other components of the model, we perform ablation studies. We added the DLN on top of the methods mentioned in Table \ref{tab:abl_1} under the same settings provided by the original paper's authors. We report the METEOR and CIDER scores with and without the DLN on MSVD dataset. From Table \ref{tab:abl_1}, we can see that adding the DLN has significantly boosted performance.

\begin{table}[h]
    \centering
    \resizebox{\columnwidth}{!}{
    \begin{tabular}{c|c c|c c|c c}
      \hline \hline 
     Models & RL &  & MRT & & DLN &    \\
     &  M & C  & M & C & M & C \\ 
     \hline 
     SA-LSTM \cite{salstm_ref} & 32.1 & 76.7 & 32.3 & 76.1  & 33.1 & 77.2 \\
     RecNet \cite{recnet_ref} & 34.3 & 81.1 & 34.1 & 80.7 & 34.4 & 81.3  \\
     M3 \cite{m3_ref} & 33.7 & - & 33.4 & - & 34.9 & -  \\
     PickNet \cite{picknet_ref} & 33.5 & 77.8 & 33.3 & 77.1 & 34.5 & 78.7  \\
     MARN \cite{marn_ref} & 35.4 & 93.5 & 35.0 & 92.5 & 35.7 & 93.4  \\
     \hline \hline
    \end{tabular}}
    \caption{Ablation studies on  MSVD benchmark. M and C denote METEOR and CIDER respectively.}
    \label{tab:abl_2}
\end{table}

The comparison of the performance of the DLN with its competitors on direct metric training is shown in Table \ref{tab:abl_2}. The experiments are done on the above-mentioned methods under the original settings for a fair comparison. Table \ref{tab:abl_2} shows that our method outperforms its other counterparts. We report METEOR and CIDER scores for all the comparisons since these two are the most important metric to evaluate captioning tasks.

\subsection{Study on the training of the DLN. } 
\begin{figure}[h]
\includegraphics[width=0.9\columnwidth,height=230px]{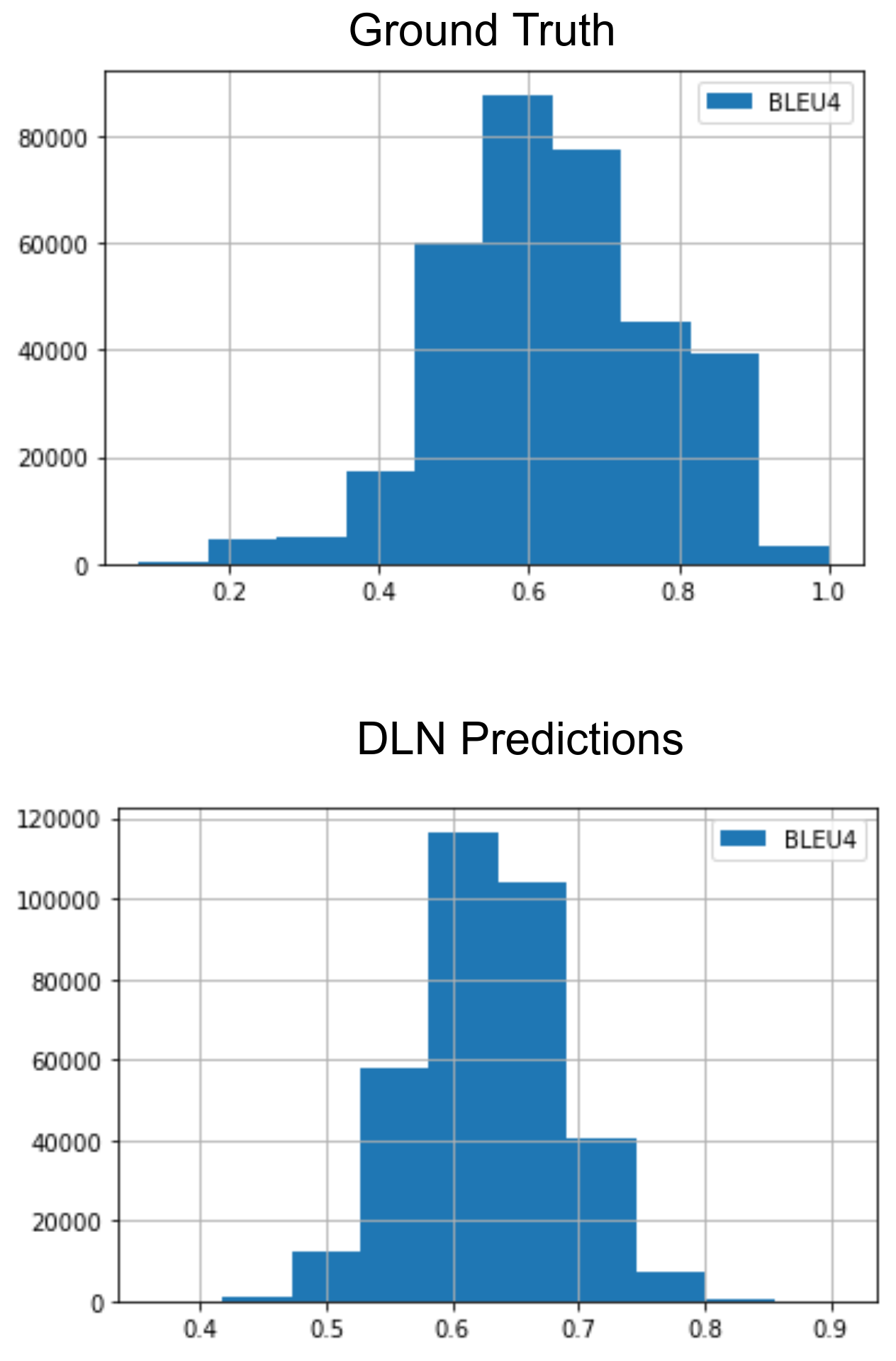}
\caption{Comparison of BLEU-4 Histograms: ground truth vs model prediction.}
\label{fighist}
\end{figure}

The training of the DLN is performed to predict $BLEU$, $METEOR$, and $CIDER$. When it comes to $ROUGE$ modeling, DLN is not as effective as other measures. Also, the signal from $ROUGE$ is not helpful to boost the model performance. The novel idea of the DLN is proposed in this paper, so no benchmark results are available for this task. Hence, the qualitative analysis is performed by comparing histograms of the ground truth and the predicted values on the test set, as shown in Fig.\ref{fighist}. We have given the $BLEU$ results, whereas the $METEOR$ and $CIDER$ stage-1 training outcomes are also similar.

\subsection{Qualitative Results }
\begin{figure}[h]
    \includegraphics[width=\columnwidth]{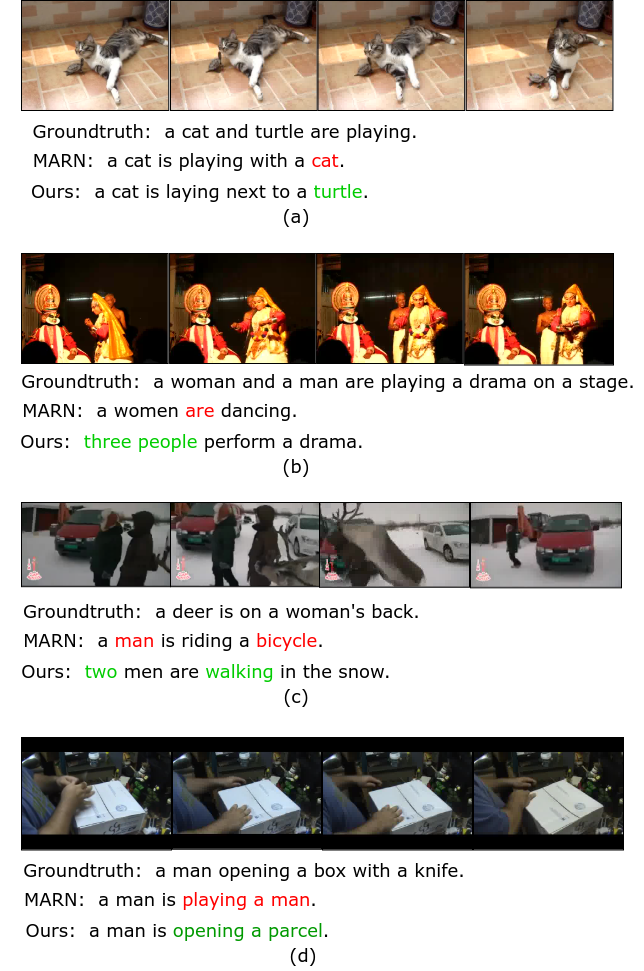}
    \caption{Qualitative comparison of Captions generated by our model and MARN\cite{marn_ref}.}
    \label{fig:qualitative_result}
\end{figure}

The Fig.\ref{fig:qualitative_result} shows the captions generated by our model and MARN \cite{marn_ref}. From the figure, we can see that our proposed model performs better than MARN \cite{marn_ref} in detecting objects and actions. Also, the captions generated by our model are more grammatically sound.

\section{Conclusion} This work addresses the training signal evaluation metric alignment mismatch problem of existing video captioning models and proposes a dynamic loss network (DLN), which models the evaluation metric under consideration. The training is performed in two stages, and the experimental results on the benchmark datasets show superior performance than current state-of-the-art models. Also, our approach shows better performance than other existing non-differentiable training strategies for video captioning and can be easily adaptable to similar tasks. Future studies could investigate the effectiveness of our method on other tasks such as image captioning and machine translation.






\bibliographystyle{IEEEtran}
\bibliography{reference}
%



\end{document}